\documentclass[runningheads]{llncs}
\usepackage[T1]{fontenc}
\usepackage{graphicx}
\usepackage{amsmath,amssymb,amsfonts}
\usepackage{cite}
\usepackage{orcidlink}
\usepackage[symbol]{footmisc}

\def\BibTeX{{\rm B\kern-.05em{\sc i\kern-.025em b}\kern-.08em
    T\kern-.1667em\lower.7ex\hbox{E}\kern-.125emX}}

\begin{document}
\title{From Latent to Engine Manifolds: Analyzing ImageBind's Multimodal Embedding Space}
\titlerunning{Analyzing ImageBind's Multimodal Embedding Space}
%
\author{Andrew Hamara \and
Pablo Rivas~\orcidlink{0000-0002-8690-0987}}
\authorrunning{A. Hamara and P. Rivas}
%
\institute{Department of Computer Science, Baylor University, Texas, USA
\email{\{Andrew\_Hamara1,Pablo\_Rivas\}@Baylor.edu}}
\maketitle              
\begin{abstract}
This study investigates ImageBind's ability to generate meaningful fused multimodal embeddings for online auto parts listings. We propose a simplistic embedding fusion workflow that aims to capture the overlapping information of image/text pairs, ultimately combining the semantics of a post into a joint embedding. After storing such fused embeddings in a vector database, we experiment with dimensionality reduction and provide empirical evidence to convey the semantic quality of the joint embeddings by clustering and examining the posts nearest to each cluster centroid. Additionally, our initial findings with ImageBind’s emergent zero-shot cross-modal retrieval suggest that pure audio embeddings can correlate with semantically similar marketplace listings, indicating potential avenues for future research.

\end{abstract}
\section{Introduction}
\begin{figure}[b!]
    \centering
    \includegraphics[width=\textwidth]{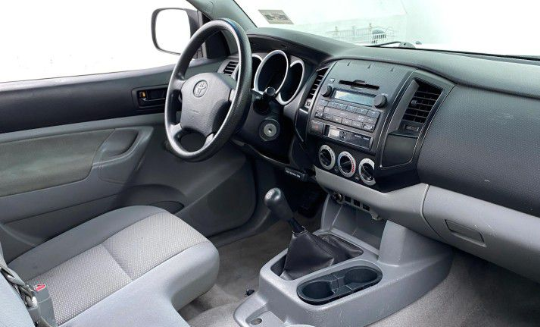}
    \includegraphics[width=0.8\textwidth]{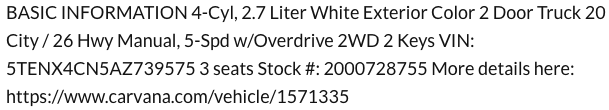}
    \caption{Example of a vehicle posted for sale online. It has a textual description and an image out of many.}
    \label{fig:sample}
\end{figure}

The growth of online marketplaces has yielded a massive dataset of image/text pairs that loosely describe the same item but often contain noise (e.g. contact information, payment preference, URLs), as shown in Fig.~\ref{fig:sample}. Concurrently, the introduction of the transformer architecture \cite{attention} has led to significant advancements \cite{elmo, xbert, visiontransformer, dall-e, openai-gpt-3} in deep learning feature representation. More specifically, models trained on large corpora have proven \cite{fernandes2018supervised, littmann2021embeddings, openai-gpt-3} to capture semantics by encoding inputs into embedding vectors (embeddings).

Research of such models has shifted toward multimodality \cite{clip, girdhar2023imagebind, llava, flamingo, chameleon, zhang2022contrastivelearningmedicalvisual}, encoding several modalities into the same embedding space, often using a contrastive loss function. Studies on these models \cite{xbert, girdhar2023imagebind} have demonstrated that robust embedding spaces preserve addition and subtraction, and average word embeddings (AWEs) have been widely adopted \cite{coates2018frustratinglyeasymetaembedding, yt_avgwordembeddings} to represent the general semantics of a sentence, particularly when weighted \cite{weighted_avg_word, arefyev2018doeswordweighweighting, de2016representation, arora2017simple}. Exploration and discussion of averaging cross-modal embeddings, however, remains limited in the literature, even within sections of embedding arithmetic\cite{girdhar2023imagebind}.

Our research focuses on leveraging ImageBind \cite{girdhar2023imagebind} to fuse image/text embeddings into a joint representation to capture overlapping information between the modalities. The main contribution of this research is empirical evidence that averaging cross-modal embeddings is an effective, straightforward mechanism for multimodal feature representation. Finally, we complement existing literature \cite{musicrecommendation, yt_avgwordembeddings} regarding averaging embeddings and embedding arithmetic and provide insights into the types of auto parts listings dominating the marketplace.

In an effort to make this paper self-contained, our discussion begins in Section 2 with an overview of ImageBind and its overall architecture. We then introduce our dataset and workflow for utilizing ImageBind in Section 3, including embedding fusion, storage, and clustering. In Section 4, the resulting embeddings and clusters are analyzed, visualized, and defined, with empirical evidence provided to convey their significance. Section 4 also includes a brief experiment that shows promising results regarding ImageBind's cross-modal retrieval. We conclude in Section 5 with a discussion about how this workflow can be adopted in practice and its implications for our future research.

\section{ImageBind Overview}

This section summarizes ImageBind's architecture as developed by its authors, aiming to briefly capture the essence of their design and contextualize its utility in our research.

\subsection{Modalities}
ImageBind combines six different data types: text, image/video, thermal, depth, audio, and IMU into a unified embedding space. With images as an anchor, this space allows for representing semantic meaning across different modalities, even those that are not typically paired together in datasets. We utilize the text, image, and audio modalities in our study. While we acknowledge that our results could potentially be replicated with a simpler model (e.g. \cite{audioclip}), we chose ImageBind to maintain as much flexibility as possible in future research.

\subsection{Encoders}
ImageBind uses separate encoders for all six modalities. The Vision Transformer (ViT) is used for images and video. Audio, depth, and thermal data are also processed with ViT, where audio is first converted into spectrograms, and depth and thermal data are treated as single-channel images. Finally, they follow the text encoder design from CLIP \cite{clip}. A linear projection head is added to each encoder, yielding a fixed size \textit{d}-dimensional embedding across all modalities.

\subsection{Training}
ImageBind was trained to create a joint embedding space using pairs of modalities, specifically images \(I\) and another modality \(M\), leveraging large-scale web datasets covering a broad semantic spectrum. It also incorporates self-supervised pairings of images with other modalities including audio, depth, thermal, and Inertial Measurement Unit (IMU) data.

For each image \( I_i \) and its corresponding observation in another modality 
\( M_i \), ImageBind generates normalized embeddings $q_i$ = $f$(\(I_i)\) and $k_i$ = $g$(\(M_i)\), respectively, where $f$ and $g$ are deep networks. The encoders are then optimized via the InfoNCE \cite{infonce} loss function:
\begin{equation}
L_{I,M} = -\log \left( \frac{e^{q_i^{\,\intercal} k_i / \tau}}{e^{q_i^{\,\intercal} k_i / \tau} + \sum_{j \neq i} e^{q_i^{\,\intercal} k_j / \tau}} \right), \nonumber   
\end{equation}
where $\tau$ is a scalar affecting the softmax distribution, and $j$ refers to an unrelated observation.  InfoNCE aims to maximize the similarity between the related embeddings (positives) $q_i$ and $k_i$ and minimize the similarity between $q_i$ and a set of unrelated observations (negatives). 

ImageBind demonstrates \textit{emergent} behavior where it aligns embeddings of two different non-image modalities $M1, M2$ that were not directly paired during training but were both paired with images $I$. Using the language of the authors, $I$ is said to \textit{bind} modalities $M1, M2$ in the embedding space, allowing for cross-modal retrieval between them. Since ImageBind was not explicitly trained for such retrieval across $M1, M2$, this is said to be \textit{emergent} cross-modal retrieval. 

\section{Methodology}

Our dataset comprises text and image data extracted from online C2C auto parts listings featuring textual descriptions and at least one corresponding image of the item(s) for sale. In cases where listings contain multiple images, we assume that each image is equally relevant to the listing. Additionally, we consider the images and text to hold equal weight in representing the post. We processed a total of 50k posts with a total of 220k images. Since the dataset contains personally identifiable information, it will remain closed to ensure privacy.

\subsection{Embedding Fusion}

Denote $e_{\text{image}}^{(i)}$ as the embedding vector of the $i$-th image of a post, with $i$ ranging from 1-$n$, and $n$ being the number of images within the post. The text content of the post is represented by the embedding vector $e_{\text{text}}$. Taking advantage of ImageBind's embedding arithmetic, we average embeddings to preserve general semantic similarities \cite{briand2021semipersonalized, DNNYR2016}. We calculate a mean image embedding as the arithmetic mean of the individual image embeddings, i.e., $e_{\text{avg\_image}} = \frac{1}{n} \sum_{i=1}^{n} e_{\text{image}}^{(i)}$.

To ensure a balanced contribution of the text embedding and mean image embedding in the fused result, the combined embedding is scaled by a factor of 0.5 \cite{girdhar2023imagebind}. Thus, the fused multimodal embedding is formally expressed as:
\[
e_{\text{multimodal}} = \frac{1}{2} \left( \frac{1}{n} \sum_{i=1}^{n} e_{\text{image}}^{(i)} + e_{\text{text}} \right).
\]

\begin{figure}[b!]
    \centering
    \includegraphics[width=1.0\columnwidth, height=.195\textheight]{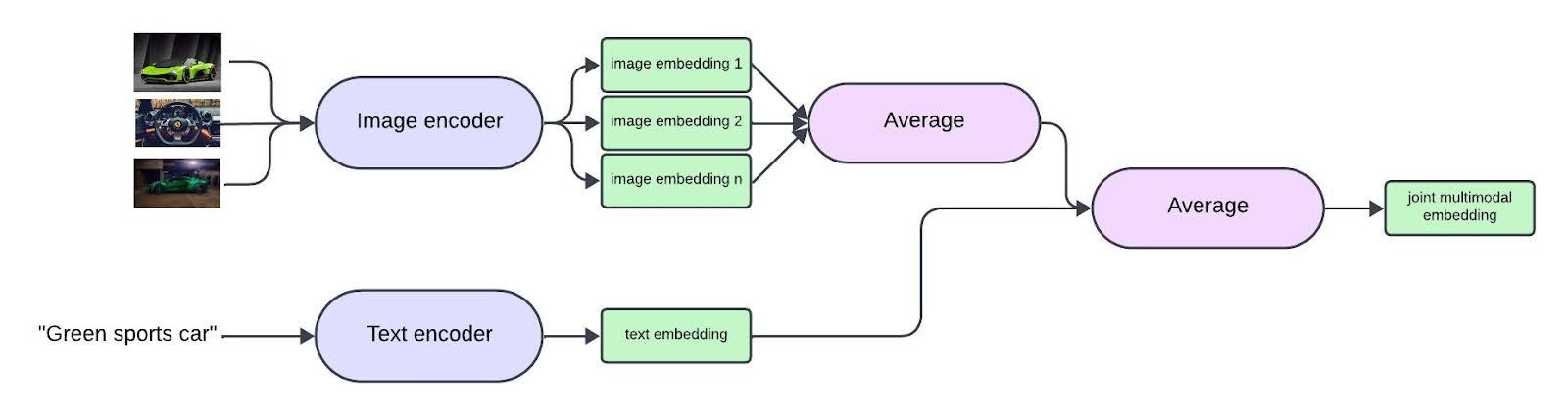}
    \caption{A "frustratingly" \cite{frustratingly_simple} simple workflow for creating joint, multimodal embeddings.}
    \label{fig:workflow}
\end{figure}

This workflow is depicted in Fig.~\ref{fig:workflow} for a single post. The process is applied to each listing and the resulting fused embeddings are stored in a vector database. 

\subsection{Clustering}

After storing the fused embeddings in a vector database, we apply Principal Component Analysis (PCA) to reduce their dimensionality, accelerating the processes of clustering and retrieval. We explore dimensionality reductions to 8, 16, 32, 64, and 128 dimensions, running $k$-means clustering on each reduced set of embeddings. The resulting centroids are evaluated against the original high-dimensional embeddings using the Silhouette score \cite{ROUSSEEUW198753}, Calinski-Harabasz index \cite{calinskiharabasz1974}, and Davies-Bouldin index \cite{daviesbouldin1979}. As shown in Table~\ref{tab:accents}, our results indicate that reducing to 32 dimensions is marginally optimal for this dataset, assuming equal weight of the metrics.

\begin{table}[h!]
  \centering
  \caption{Reduced embedding dimensions and their corresponding index scores. Calinski-Harabasz and Davies-Bouldin are abbreviated as C-H and D-B, respectively.}
  \begin{tabular}{rrrr}
    \hline
    \textbf{Dim.} & \textbf{Silhouette} & \textbf{C-H} & \textbf{D-B} \\
    \hline
    8      & .3726 & 672.1 & 4.945 \\
    16     & .3799 & 696.5 & 4.353 \\
    32     & .3819 & 709.1 & 4.043 \\
    64     & .3810 & 709.3 & 4.091\\
    128    & .3816 & 710.2 & 4.134 \\
    \hline
  \end{tabular}
  \label{tab:accents}
\end{table}

Once reduced, we cluster the fused embeddings with $k$-means \cite{lloyd1982least} \cite{scikit-learn}, chosen for its simplicity and broad application. We follow \cite{garg2018supervising} in choosing $k$ (ranging from 2 to 20 for visualization purposes) by conducting ten $k$-means iterations with different random states for each $k$ and selecting the $k$ with the highest average Silhouette score among all iterations. This process revealed that $k=20$ provided the best visual\footnote{Following this process ranging from 2 to 250, i.e., with no color limitations, revealed $k=37$ to be optimal.} structuring for our data.

\section{Results and Discussion}

We apply Uniform Manifold Approximation and Projection (UMAP) \cite{mcinnes2020umap} to the original high-dimensional embeddings to reduce them to 2D for visualization, using the labels from the 32-dimensional clusters for colorization. This visual representation aids in the subsequent analysis of the clusters, where we explore the patterns that define the structure of our data.

\subsection{Cluster Analysis}

\begin{figure}[h!]
  \centering
  \includegraphics[width=\columnwidth, trim=2cm 1.5cm 3.75cm 2.25cm, clip]{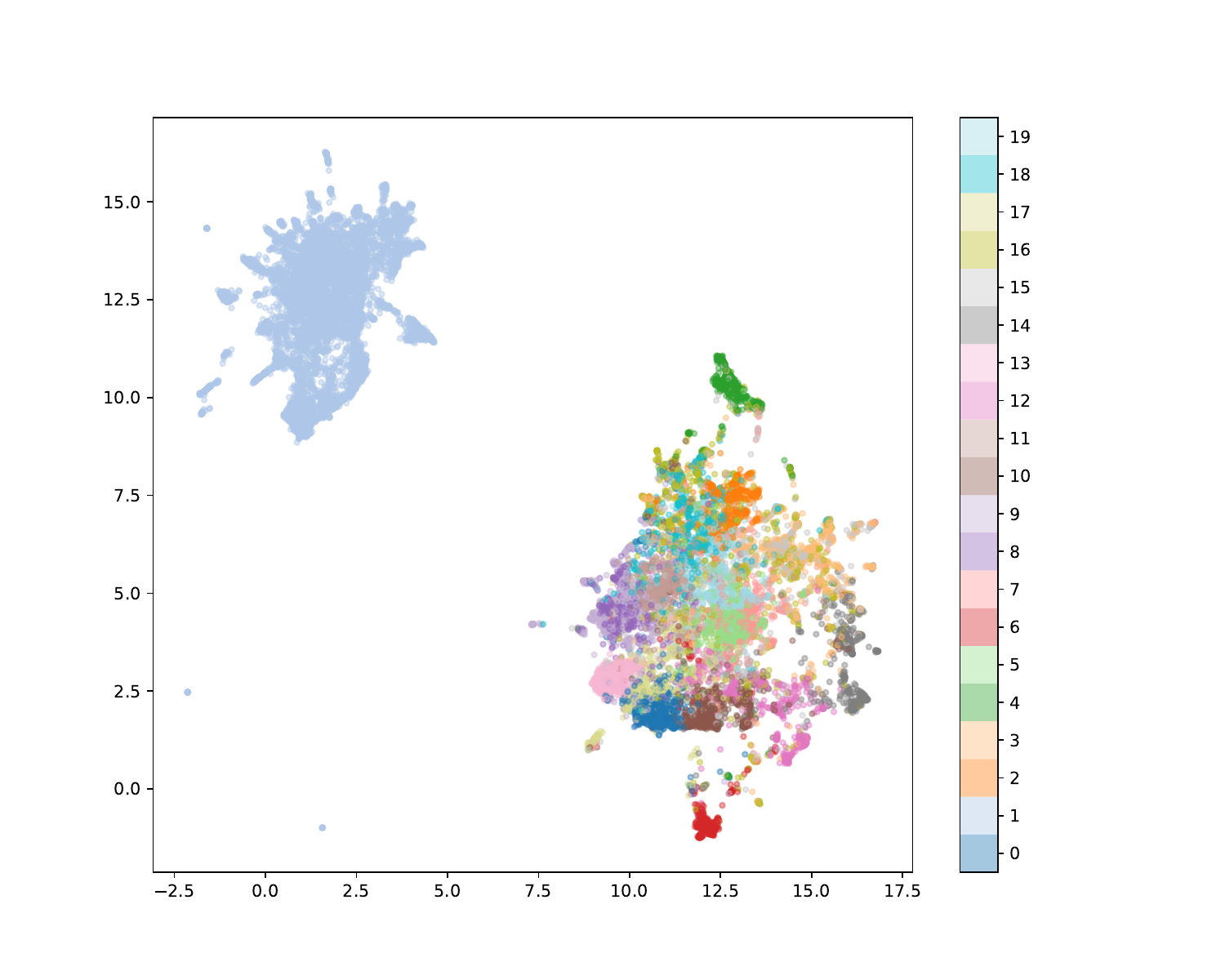}
  \caption{UMAP visualization of the high-dimensional embeddings, with clusters colorized based on 32-dimensional $k$-means results.}
  \label{fig:umap-clusters}
\end{figure}

The clusters revealed from our analysis showcase distinct patterns and group characteristics within the dataset. Most notably, cluster 0, depicted in the upper-left corner of Fig.~\ref{fig:umap-clusters}, is outlying.

\begin{figure}[h!]
    \centering
    \includegraphics[width=\textwidth]{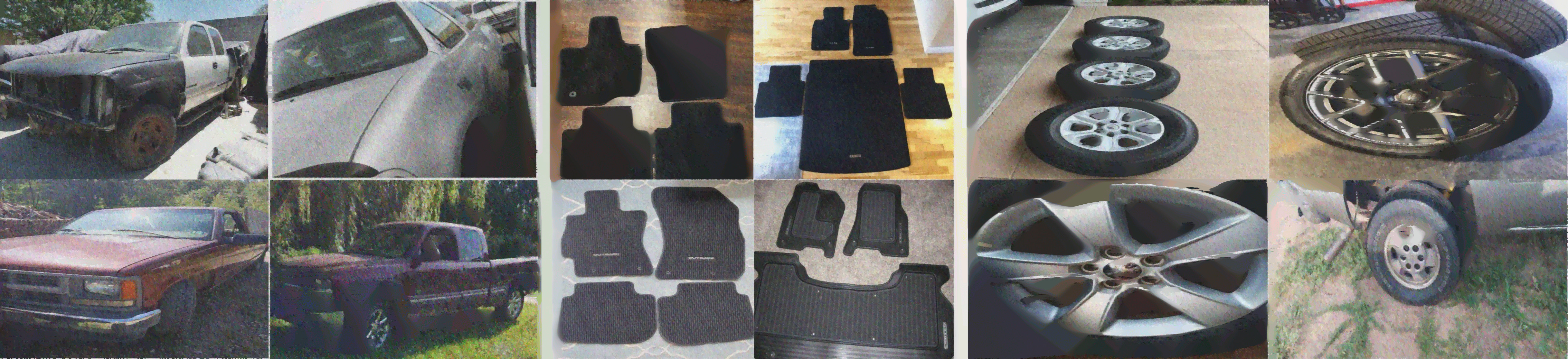}
    \caption{Images of posts near their respective $k$-means centroid, grouped by cluster.}
    \label{fig:empirical}
\end{figure}

To understand each $k$-means cluster's characteristics, we identified the ten nearest posts by Euclidean distance to each cluster's centroid. The close alignment of these nearest neighbors with their respective cluster centers, such as those shown in Fig.~\ref{fig:empirical}, affirmed the validity of the clustering outcome, suggesting that the clusters capture distinct, meaningful groupings within the data.

Furthermore, we found that the most representative posts from Cluster 0, identified as the outlying cluster, were instances of 'parting out.' In these cases, users list a myriad of parts for sale, accompanied by textual descriptions and images of the corresponding parts. These listings are distinct from others in that there is typically little alignment between the text and images; each sentence may describe one of many images, and each image may only correspond to a subset of the entire textual description. Notably, more than 30\% of all embeddings were grouped in this cluster, suggesting that such listings are highly prevalent in the online auto parts marketplace.

Clusters 1-19 correspond to a specific auto part category, highlighting the effectiveness of ImageBind's embedding space and our simplistic fusion workflow. We identified clusters for components such as 'intakes, manifolds, engines' (Cluster 1) and 'panels, bumpers, door parts' (Cluster 3), as well as vehicle types and unique fittings like 'sedans' (Clusters 5 and 6), 'truck beds and bed shells' (Cluster 8), and 'tires/wheels' (Cluster 13). More specific items, including 'trailer hitches' (Cluster 14) and 'lights' (Cluster 15), demonstrate ImageBind's robust embedding space and efficacy in classifying C2C auto parts listings.

\subsection{Cross-Modal Retrieval}

\begin{figure}
    \centering
    \includegraphics[width=\textwidth]{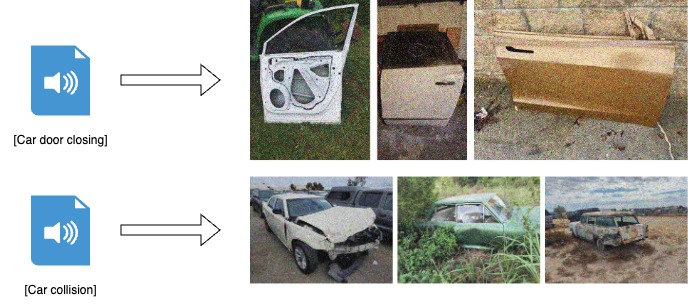}
    \caption{Retrieval of fused listing embeddings via semantically similar audio embeddings.}
    \label{fig:audio}
\end{figure}

Further exploring ImageBind's set of modalities, we generate pure audio embeddings for sounds related to auto parts, e.g. car doors closing, car collisions, or engines revving. We then applied $k$-NN to map these audio embeddings to their nearest counterparts in our database of fused listing embeddings. ImageBind successfully aligned a wide range of such sounds with their corresponding image/text counterparts, as shown Fig.~\ref{fig:audio}, demonstrating the model's ability to create meaningful cross-modal associations and potentially have applications in audio or video based recommendation.

\section{Conclusions}
Our analysis demonstrates that ImageBind is a powerful tool for interpreting online C2C auto parts listings. By examining clusters and the corresponding $k$-NN posts for each $k$-means centroid, we found that meaningful fused image/text embeddings could be created through simple averaging of the corresponding text/image embeddings for each listing. This research not only proves the quality of ImageBind's embedding space and the efficacy of its embedding arithmetic but also indicates potential for applications in filtering and recommendation systems within C2C marketplaces. Our exploration of audio embeddings further proves ImageBind's cross-modal capabilities, presenting a promising avenue for future research.

\begin{credits}
\subsubsection{\ackname} Part of this work was funded by the National Science Foundation under grants CNS-2210091 and CNS-2136961.

\subsubsection{\discintname}
The authors have no competing interests to declare that are relevant to the content of this article.
\end{credits}
%
%
%
%
\bibliographystyle{splncs04}
\bibliography{refs}

\begin{thebibliography}{10}
\providecommand{\url}[1]{\texttt{#1}}
\providecommand{\urlprefix}{URL }
\providecommand{\doi}[1]{https://doi.org/#1}

\bibitem{flamingo}
Alayrac, J.B., Donahue, J., Luc, P., Miech, A., Barr, I., Hasson, Y., Lenc, K., Mensch, A., Millican, K., Reynolds, M., Ring, R., Rutherford, E., Cabi, S., Han, T., Gong, Z., Samangooei, S., Monteiro, M., Menick, J., Borgeaud, S., Brock, A., Nematzadeh, A., Sharifzadeh, S., Binkowski, M., Barreira, R., Vinyals, O., Zisserman, A., Simonyan, K.: Flamingo: a visual language model for few-shot learning (2022), \url{https://arxiv.org/abs/2204.14198}

\bibitem{arefyev2018doeswordweighweighting}
Arefyev, N., Ermolaev, P., Panchenko, A.: How much does a word weigh? weighting word embeddings for word sense induction (2018), \url{https://arxiv.org/abs/1805.09209}

\bibitem{arora2017simple}
Arora, S., Liang, Y., Ma, T.: A simple but tough-to-beat baseline for sentence embeddings. In: International conference on learning representations (2017)

\bibitem{briand2021semipersonalized}
Briand, L., Salha-Galvan, G., Bendada, W., Morlon, M., Tran, V.A.: A semi-personalized system for user cold start recommendation on music streaming apps (2021)

\bibitem{openai-gpt-3}
Brown, T.B., Mann, B., Ryder, N., Subbiah, M., Kaplan, J., Dhariwal, P., Neelakantan, A., Shyam, P., Sastry, G., Askell, A., Agarwal, S., Herbert-Voss, A., Krueger, G., Henighan, T., Child, R., Ramesh, A., Ziegler, D.M., Wu, J., Winter, C., Hesse, C., Chen, M., Sigler, E., Litwin, M., Gray, S., Chess, B., Clark, J., Berner, C., McCandlish, S., Radford, A., Sutskever, I., Amodei, D.: Language models are few-shot learners (2020), \url{https://arxiv.org/abs/2005.14165}

\bibitem{calinskiharabasz1974}
Caliński, T., Harabasz, J.: A dendrite method for cluster analysis. Communications in Statistics  \textbf{3}(1),  1--27 (1974). \doi{10.1080/03610927408827101}, \url{https://www.tandfonline.com/doi/abs/10.1080/03610927408827101}

\bibitem{coates2018frustratinglyeasymetaembedding}
Coates, J., Bollegala, D.: Frustratingly easy meta-embedding -- computing meta-embeddings by averaging source word embeddings (2018), \url{https://arxiv.org/abs/1804.05262}

\bibitem{frustratingly_simple}
Coates, J., Bollegala, D.: Frustratingly easy meta-embedding -- computing meta-embeddings by averaging source word embeddings (2018), \url{https://arxiv.org/abs/1804.05262}

\bibitem{DNNYR2016}
Covington, P., Adams, J., Sargin, E.: Deep neural networks for youtube recommendations. In: Proceedings of the 10th ACM Conference on Recommender Systems. p. 191–198. RecSys '16, Association for Computing Machinery, New York, NY, USA (2016). \doi{10.1145/2959100.2959190}, \url{https://doi.org/10.1145/2959100.2959190}

\bibitem{daviesbouldin1979}
Davies, D.L., Bouldin, D.W.: A cluster separation measure. IEEE Transactions on Pattern Analysis and Machine Intelligence  \textbf{PAMI-1}(2),  224--227 (1979). \doi{10.1109/TPAMI.1979.4766909}

\bibitem{de2016representation}
De~Boom, C., Van~Canneyt, S., Demeester, T., Dhoedt, B.: Representation learning for very short texts using weighted word embedding aggregation. Pattern Recognition Letters  \textbf{80},  150--156 (2016)

\bibitem{xbert}
Devlin, J., Chang, M.W., Lee, K., Toutanova, K.: {BERT}: Pre-training of deep bidirectional transformers for language understanding. In: Burstein, J., Doran, C., Solorio, T. (eds.) Proceedings of the 2019 Conference of the North {A}merican Chapter of the Association for Computational Linguistics: Human Language Technologies, Volume 1 (Long and Short Papers). pp. 4171--4186. Association for Computational Linguistics, Minneapolis, Minnesota (Jun 2019). \doi{10.18653/v1/N19-1423}, \url{https://aclanthology.org/N19-1423}

\bibitem{visiontransformer}
Dosovitskiy, A., Beyer, L., Kolesnikov, A., Weissenborn, D., Zhai, X., Unterthiner, T., Dehghani, M., Minderer, M., Heigold, G., Gelly, S., Uszkoreit, J., Houlsby, N.: An image is worth 16x16 words: Transformers for image recognition at scale (2021), \url{https://arxiv.org/abs/2010.11929}

\bibitem{weighted_avg_word}
Elsaadawy, A., Torki, M., Ei-Makky, N.: A text classifier using weighted average word embedding. In: 2018 International Japan-Africa Conference on Electronics, Communications and Computations (JAC-ECC). pp. 151--154 (2018). \doi{10.1109/JEC-ECC.2018.8679539}

\bibitem{fernandes2018supervised}
Fernandes, K., Chicco, D., Cardoso, J.S., Fernandes, J.: Supervised deep learning embeddings for the prediction of cervical cancer diagnosis. PeerJ Computer Science  \textbf{4}, ~e154 (2018)

\bibitem{garg2018supervising}
Garg, V.K., Kalai, A.: Supervising unsupervised learning (2018)

\bibitem{girdhar2023imagebind}
Girdhar, R., El-Nouby, A., Liu, Z., Singh, M., Alwala, K.V., Joulin, A., Misra, I.: Imagebind: One embedding space to bind them all (2023)

\bibitem{audioclip}
Guzhov, A., Raue, F., Hees, J., Dengel, A.: Audioclip: Extending clip to image, text and audio (2021), \url{https://arxiv.org/abs/2106.13043}

\bibitem{musicrecommendation}
Hansen, C., Hansen, C., Maystre, L., Mehrotra, R., Brost, B., Tomasi, F., Lalmas, M.: Contextual and sequential user embeddings for large-scale music recommendation. In: Proceedings of the 14th ACM Conference on Recommender Systems. p. 53–62. RecSys '20, Association for Computing Machinery, New York, NY, USA (2020). \doi{10.1145/3383313.3412248}, \url{https://doi.org/10.1145/3383313.3412248}

\bibitem{littmann2021embeddings}
Littmann, M., Heinzinger, M., Dallago, C., Olenyi, T., Rost, B.: Embeddings from deep learning transfer go annotations beyond homology. Scientific reports  \textbf{11}(1), ~1160 (2021)

\bibitem{llava}
Liu, H., Li, C., Wu, Q., Lee, Y.J.: Visual instruction tuning (2023), \url{https://arxiv.org/abs/2304.08485}

\bibitem{lloyd1982least}
Lloyd, S.: Least squares quantization in pcm. IEEE transactions on information theory  \textbf{28}(2),  129--137 (1982)

\bibitem{mcinnes2020umap}
McInnes, L., Healy, J., Melville, J.: Umap: Uniform manifold approximation and projection for dimension reduction (2020)

\bibitem{infonce}
van~den Oord, A., Li, Y., Vinyals, O.: Representation learning with contrastive predictive coding (2019), \url{https://arxiv.org/abs/1807.03748}

\bibitem{scikit-learn}
Pedregosa, F., Varoquaux, G., Gramfort, A., Michel, V., Thirion, B., Grisel, O., Blondel, M., Prettenhofer, P., Weiss, R., Dubourg, V., Vanderplas, J., Passos, A., Cournapeau, D., Brucher, M., Perrot, M., Duchesnay, E.: Scikit-learn: Machine learning in {P}ython. Journal of Machine Learning Research  \textbf{12},  2825--2830 (2011)

\bibitem{elmo}
Peters, M.E., Neumann, M., Iyyer, M., Gardner, M., Clark, C., Lee, K., Zettlemoyer, L.: Deep contextualized word representations. In: Walker, M., Ji, H., Stent, A. (eds.) Proceedings of the 2018 Conference of the North {A}merican Chapter of the Association for Computational Linguistics: Human Language Technologies, Volume 1 (Long Papers). pp. 2227--2237. Association for Computational Linguistics, New Orleans, Louisiana (Jun 2018). \doi{10.18653/v1/N18-1202}, \url{https://aclanthology.org/N18-1202}

\bibitem{clip}
Radford, A., Kim, J.W., Hallacy, C., Ramesh, A., Goh, G., Agarwal, S., Sastry, G., Askell, A., Mishkin, P., Clark, J., Krueger, G., Sutskever, I.: Learning transferable visual models from natural language supervision (2021), \url{https://arxiv.org/abs/2103.00020}

\bibitem{dall-e}
Ramesh, A., Dhariwal, P., Nichol, A., Chu, C., Chen, M.: Hierarchical text-conditional image generation with clip latents (2022), \url{https://arxiv.org/abs/2204.06125}

\bibitem{ROUSSEEUW198753}
Rousseeuw, P.J.: Silhouettes: A graphical aid to the interpretation and validation of cluster analysis. Journal of Computational and Applied Mathematics  \textbf{20},  53--65 (1987). \doi{https://doi.org/10.1016/0377-0427(87)90125-7}, \url{https://www.sciencedirect.com/science/article/pii/0377042787901257}

\bibitem{yt_avgwordembeddings}
Savigny, J., Purwarianti, A.: Emotion classification on youtube comments using word embedding. In: 2017 International Conference on Advanced Informatics, Concepts, Theory, and Applications (ICAICTA). pp.~1--5 (2017). \doi{10.1109/ICAICTA.2017.8090986}

\bibitem{chameleon}
Team, C.: Chameleon: Mixed-modal early-fusion foundation models (2024), \url{https://arxiv.org/abs/2405.09818}

\bibitem{attention}
Vaswani, A., Shazeer, N., Parmar, N., Uszkoreit, J., Jones, L., Gomez, A.N., Kaiser, L., Polosukhin, I.: Attention is all you need. In: Proceedings of the 31st International Conference on Neural Information Processing Systems. p. 6000–6010. NIPS'17, Curran Associates Inc., Red Hook, NY, USA (2017)

\bibitem{zhang2022contrastivelearningmedicalvisual}
Zhang, Y., Jiang, H., Miura, Y., Manning, C.D., Langlotz, C.P.: Contrastive learning of medical visual representations from paired images and text (2022), \url{https://arxiv.org/abs/2010.00747}

\end{thebibliography}

\end{document}